\documentclass[lettersize,journal]{IEEEtran}
\usepackage{amsmath,amsfonts}
\usepackage{algorithmic}
\usepackage{algorithm}
\usepackage{array}
\usepackage[caption=false,font=normalsize,labelfont=sf,textfont=sf]{subfig}
\usepackage{textcomp}
\usepackage{url}
\usepackage{verbatim}
\usepackage{graphicx}
\usepackage{cite}

\usepackage{hyperref}

\usepackage{cleveref}
\usepackage{booktabs}
\usepackage{adjustbox}
\usepackage{balance}

\begin{document}

\title{The Neglected Baseline in Model Interpretation}

\author{Yongjin Cui, Xiaohui Fan
\thanks{Yongjin Cui, Zhejiang University, cuiyongjin@zju.edu.cn; Xiaohui Fan, Zhejiang University, fanxh@zju.edu.cn}
}



\maketitle

\begin{abstract}

  We observe that existing model interpretation methods generally ignore the baseline, and such neglect often results in imprecise or even incorrect interpretation. In this paper, we reformulate the task of model interpretation and the interpretation principles for model interpretation results to demonstrate the importance of the baseline.  We further unify gradient-based methods, Integrated Gradients (IG) methods, and Taylor expansion, clarifying the connections among them and explicitly identifying the baseline for each method. On this basis, we analyze the flaws and errors in related model interpretation methods (IG, LayerCAM, ODAM, Difference Map). We advocate evaluating the quality of model interpretation results precisely through the attribution error between the attribution result and the attribution target, rather than adopting flawed evaluation methods, such as those based on  marginal-effect or the assumption of perfect model performance. We revise IG and develope a model interpretation method with a clear and reasonable baseline, achieving better results. Our method supports model interpretation based on features from any layer. Interpretation based on features from different layers are all reasonable, and the differences among these results reflect varying degrees of feature extraction at different feature extraction stages.
  
  \textit{This work has been submitted to the IEEE for possible publication. Copyright may be transferred without notice, after which this version may no longer be accessible.}

\end{abstract}

\begin{IEEEkeywords}
Baseline, model interpretation.
\end{IEEEkeywords}

\section{Introduction}
  \IEEEPARstart{W}{ith} the widespread application of deep learning models in fields such as computer vision and natural language processing, the ``black-box" nature of models has emerged as a significant obstacle to their further deployment in critical domains, including medical diagnostics and financial risk control. Model interpretation refers to interpret or present the rationale behind a model's decisions. It not only relates to the transparency of the model but also directly influences people's trust in and acceptance of the model. The interpretation results of models also facilitate accurate evaluation of model capabilities and guide model optimization. In recent years, academic and industrial research on model interpretation has deepened, leading to the development of various interpretation methods and technical approaches, which has been used to identify failure modes \cite{odam22,odam33}, debug models\cite{odam44} and establish appropriate users' confidence about models \cite{51}, instruct the knowledge distillation of object detection\cite{ODAM2}, and distinguish the duplicate detected objects\cite{ODAM2}.

    Current model interpretation methods can be broadly categorized into two types and others as discussed in \Cref{Related Works}. Whether in terms of the computational process or the interpretation of model interpretation results, existing methods generally neglect the baseline. Among these methods, Integrated Gradients (IG) \cite{61} involves a baseline, but it still has certain drawbacks. Moreover, current model interpretation methods based on IG also also contain serious errors \cite{cui2026IntegratedGradients}.

  Neglecting the baseline can render an method theoretically imprecise or even erroneous, while also undermining the rigor of result interpretation. The baseline serves as the foundation for our evaluation of things. For instance, when assessing the speed of a car, the default baseline is a stationary object. Without this baseline, any evaluation of speed becomes meaningless. Similarly, if the baseline setting of an method is inherently vague or uncertain, the interpretation method itself becomes uninterpretable. In interpreting model interpretation results, the absence of a baseline also renders the interpretation meaningless.

  Gradient is widely applied in model interpretation. However, methods that directly utilize gradient suffer from two significant drawbacks: first, they are prone to large errors, as these approaches essentially represent a linear approximation of complex nonlinear processes; second, they lack a clearly defined baseline. Methods of this type are exemplified by LayerCAM \cite{52}, ODAM \cite{ODAM1,ODAM2} and Difference Map (DM)\cite{DifferenceMap}, with ODAM essentially representing the application of LayerCAM in object detection models and DM being an extension of ODAM. ODAM successively published at ICLR 2023 and TPAMI 2024. DM published at ICCVW2025. To a certain extent, this indicates that mainstream authorities in the field have not yet recognized the severe consequences of neglecting the baseline.
  
  The introduction of Integrated Gradients (IG) \cite{61} offers the potential to address the drawbacks of gradient-based methods. However, at present, the IG method itself remains imperfect and even has certain defects. In practical applications, it still encounters numerous issues; for example, process-based methods such as TAM\cite{yuan2021explaining}, \cite{46}, and DIX\cite{47}, which all utilize IG, each contain errors \cite{cui2026IntegratedGradients}.

  Taylor expansion is an extremely important mathematical tool that represents a function, which is differentiable at a certain point (the baseline), as an infinite series (polynomial), thereby approximating the function with a polynomial. Taylor expansion inherently possesses a clear baseline, which aligns perfectly with the theme of our study. In the main text, we will unify gradient-based method, Integrated Gradients (IG) based method, and Taylor expansion, clarifying the connections among them.

  The evaluation methods currently widely adopted for model interpretation methods primarily rely on marginal-effect or the assumption of perfect model performance. These evaluation methods are very crude, and are even incorrect in some cases. Developing a better evaluation method for model interpretation results is an urgent problem that needs to be addressed.
  
  In response to the aforementioned issues, this paper reformulates the task of model interpretation and the interpretation principles for model interpretation results to demonstrate the importance of the baseline.  We further unify gradient-based methods, Integrated Gradients (IG) methods, and Taylor expansion, clarifying the connections among them and explicitly identifying the baseline for each method. On this basis, we analyze the flaws and errors in related model interpretation methods (IG, LayerCAM, ODAM, Difference Map). We advocate evaluating the quality of model interpretation results precisely through the attribution error between the attribution result and the attribution target, rather than adopting flawed evaluation methods, such as those based on  marginal-effect or the assumption of perfect model performance. We revise IG and develope a model interpretation method with a clear and reasonable baseline, achieving better results. Our method supports model interpretation based on features from any layer. Interpretation based on features from different layers are all reasonable, and the differences among these results reflect varying degrees of feature extraction at different feature extraction stages.
  
  Our contributions can be summarized as follows:
  \begin{itemize}
    \item We redefine the task of model interpretation and the principles for interpreting model interpretation results, demonstrating the importance of baselines.
    \item We unify gradient-based method, IG based method, and Taylor expansion, clarifying the connections among them, and analyze the flaws and errors in relevant methods.
    \item We elaborate on the drawbacks of existing evaluation methods for model interpretation resluts and proposed using attribution errors between to precisely evaluate the quality of model interpretation results.
    \item We revise IG and develope a model interpretation method with a clear and reasonable baseline, achieving better results. Our method supports model interpretation based on features from any layer.
  \end{itemize}

  \section{Related Works}\label{Related Works}
  
  This paper categorizes model interpretation methods into three main types: process-based, feature-based and other interpretation methods. This classification is primarily because feature-based methods are more suitable for performance evaluation using attribution errors.
  
  \subsection{Process-Based Interpretation Methods}
  
  These methods focus on the model's decision-making process. The core idea of such methods is to ``reconstruct" the model's decision-making process, enabling humans to understand how the model reasons from input to output. A typical representative of this type of model interpretation method is those related to Transformers. Chefer et al. \cite{48} introduced Generic Attention-model interpretability (GAE), which combines gradients with multi-head attention maps and then performs attention rollout. Yuan et al. \cite{yuan2021explaining} interpret the information flow inside Vision Transformers using Markov Chain (TAM). Barkan et al. \cite{47} propose Deep Integrated interpretation (DIX), which generates interpretation maps using GAE and Integrated Gradient. Chen et al. \cite{46} propose the Beyond Intuition Method (BT) based  on IG. These methods place greater emphasis on leveraging the model's attention. The overall idea is to simulate the distribution change process of model attention as it passes through each layer of the model, using the model's overall attention to different input parts as the interpretation result. 
  
  \subsection{Feature-Based Interpretation Methods}
  
  These methods focus on quantifying the direct contribution of features to the model's output. The core idea of feature-based interpretation methods is ``feature importance," which measures the contribution of each feature to the model's prediction. Most of these methods are gradient-based methods, such as CAM \cite{50}, GradCAM \cite{51}, LayerCAM \cite{52},Grad-ECLIP \cite{gradeclip} and ODAM \cite{ODAM1,ODAM2}. CAM utilizes the activations from the convolutional layers of a CNN to obtain saliency maps. The GradCAM method generalizes the CAM method for a broader spectrum of CNN architectures by introducing gradient information to avoid the need for a Global Average Pooling (GAP) layer. LayerCAM refines the weights of feature maps using gradients and integrates multiple layers of saliency maps to achieve more detailed results, as well as the primary baseline method ODAM in our paper. Grad-ECLIP is proposed to interpret Contrastive Language-Image Pre-training (CLIP). ODAM \cite{ODAM1,ODAM2} is gradient-weighted Object Detector Activation Maps (ODAM) to interpret the predictions of object detectors. ODAM is the application of LayerCAM in object detection models. In particular, it is worth noting the Integrated Gradient method (IG) \cite{61}. IG was proposed to address the issue of large errors in gradients within saturation regions. However, in our view, another noteworthy contribution is that IG can explicitly define a baseline.
  
  \subsection{Others}

  There are some relevance propagation methods and perturbation methods. Relevance propagation methods include Layer-Wise Relevance Propagation (LRP) \cite{53}, Contrastive LRP (CLRP) \cite{54}, and Softmax-Gradient-LRP (SGLRP) \cite{55}, among others. LRP propagates relevance from the predicted class backward to the input image based on Deep Taylor Decomposition (DTD) \cite{56}. CLRP improves upon LRP by comparing the predicted category signal with other category signals to interpret the model. SGLRP is a class-discriminative extension to DTD that uses the gradient of the softmax to backpropagate the relevance of the output probability to the input image. Perturbation methods include ViT-CX \cite{DBLP:conf/ijcai/Xie0CZ23} and TIS \cite{DBLP:conf/iccvw/EnglebertSNMSCV23}. ViT-CX is based on token embeddings rather than the attentions paid to them and their causal impacts on the model output. TIS is a perturbation-based interpretability method for Vision Transformers that computes a saliency map based on perturbations induced by sampling the input tokens.

  It is particularly noteworthy here that the three process-based methods—TAM, BT, and DIX—have applied IG, yet all have been proven to contain significant errors \cite{cui2026IntegratedGradients}. Additionally, Grad-ECLIP, a feature-based method, has also been proven to have significant errors \cite{cui2026debunkinggradeclipcomprehensivestudy}.

  \section{Model Interpretation and Baseline}
  
  \subsection{Re-clarification of Model Interpretation}

  Here, we are merely re-clarifying the definition of model interpretation based solely on the model interpretation tasks involved in this study, while an authoritative, rigorous, and comprehensive definition may encompass more aspects. We emphasize here the definition of model interpretability—that is, understanding and explaining the basis on which a model makes specific decisions or predictions. Decisions or predictions represent the output of the model, and the basis for these decisions comes from the input. To conduct model interpretation, one must first measure the model's input and output. When measuring or evaluating something, there must always be a baseline. For instance, when describing vehicle speed, the default baseline is a stationary object, such as a stationary road surface or building. 
  
  So, what constitutes the baseline for a model's input and output? Previous studies have largely overlooked this issue.
  
  \subsection{Baseline of Model Interpretation}\label{sec:baseline}
    
  Here, we take the object detection model DETR \cite{DBLP:conf/eccv/CarionMSUKZ20} as an example for illustration. The model takes an image as input and produces two outputs: a bounding box and a class score.
  
  Without a baseline, it is impossible to accurately describe or evaluate a thing. In many cases where a baseline is not explicitly stated, there is actually an implicit default baseline. For instance, when we describe a vehicle's speed without specifying a reference object, the default baseline is assumed to be stationary objects. Similarly, when describing the content of an image, the default reference is a completely black image with all pixel values set to zero, rather than another non-zero baseline image. Otherwise, we would first need to subtract the baseline to clearly describe the image content. When describing the coordinates of a detection bounding box, the origin of the coordinate system serves as the baseline, and when describing the class probabilities output by the model, a probability of zero is used as the baseline.
  
  It is important to note that, in the vast majority of cases, the baselines for these three elements (input, bounding box, and class score) cannot be aligned. When a completely black image is fed into the model, it is extremely challenging to ensure that the coordinates of the target bounding boxes and the class score output by the model correspond to their respective baselines. To put it plainly, even if we input a completely black image, the model will still generate corresponding detection bounding boxes and class scroes, as shown in \Cref{fig:class,fig:box,fig:vgg}. When the baseline image (a completely black image) is input, the class probabilities and coordinates output by the model do not correspond to the expected baselines. 
    
  We have noticed that object detection models (such as DETR) have one input and two outputs. When interpreting the basis for the model's inferences, we first need to select an object from the input image and the two outputs to establish a baseline. When determining the baseline based on a certain output, for instance, if we choose the coordinate origin as the baseline, we now need to establish baselines for the input and class scroe. In this case, our approach can be summarized as first determining a baseline input through the output of the origin coordinates, and then feeding this baseline image input into the model to establish the baseline for class score. However, we find this extremely challenging because it is almost impossible to determine an input based on the output of a deep learning network.
  
  An alternative approach is to use the input to establish the baseline for the entire interpretation system. When we input an image baseline (a completely black image), the obtained object bounding box coordinates and class probabilities serve as the corresponding baselines for the outputs.
  
  More importantly, both humans and models base their understanding of image content on the numerical values of image pixels. The implicit pixel baseline is 0. Therefore, it is reasonable to choose a full-zero input as the baseline for the input when interpreting the basis for the model's output.
  
  Therefore, to provide a more accurate and reasonable interpretation of the model, it is essential to establish a baseline during the model interpretation process. Given both the difficulty and the rationality of baseline determination, the baseline should be established through the input.
  
  \subsection{Correct Interpretation of Model Interpretation Results based on the Baseline}
  Please refer to \Cref{sec:interpretation}.

  \section{Gradients, Integrated Gradients and Taylor Expansion}

  \subsection{Gradients in Model Interpretation}
  
  Gradients are often used as an indicator to measure the importance of a variable. When evaluating the contribution of a certain factor to the output, a common approach is to take the gradient as the weight of that factor and compute their product as the contribution value. This approach can be traced back to attribution methods\cite{localgradients,DBLP:journals/corr/SimonyanVZ13}. Typical methods employing this method in other models include GradCAM\cite{51}, LayerCAM\cite{52}, while in the context of Transformer interpretation, representative methods are GAE\cite{48}, ODAM\cite{ODAM1,ODAM2} and DM\cite{DifferenceMap}.
  
  Although this kind of gradient application is quite widespread, we haven't found the theoretical basis for this application approach in the aforementioned papers. \textbf{It is absurd that the principle of a model-interpretation scheme is itself difficult to explain}. Researchers have been using this method all along, yet no one has explored its theoretical underpinnings. We analyze this in \Cref{sec:unificationgrad}. This is one of the contributions of this paper.
  
  \subsubsection{The Limitation of Gradients in Model Interpretation}

\begin{figure}[!t]
      \centering
      \includegraphics[width=\linewidth]{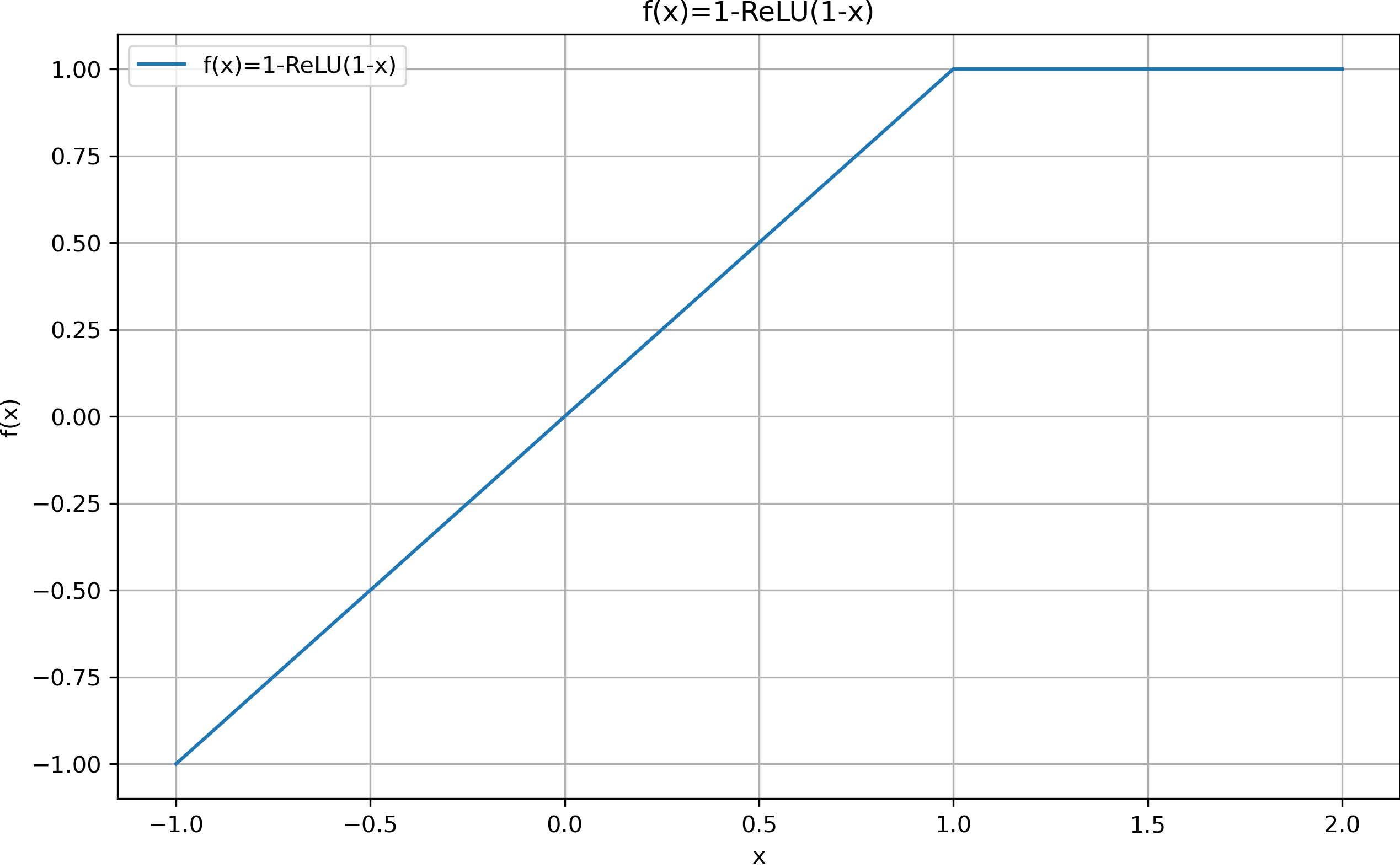} 
      \caption{ $F(x)=1-ReLU(1-x)$}
      \label{fig:relu}
  \end{figure}

  Gradient-based methods are limited because they are essentially linear approximations of complex nonlinear processes(\Cref{sec:unificationgrad}), resulting in large errors.There are even cases where it completely fails to correctly reflect the magnitude of the input's contribution. \Cref{fig:relu} shows the case of $F(x)=1-\text{ReLU}(1-x)$ presented in the IG paper \cite{61}. Taking $x=0$ as the baseline (with a baseline contribution of 0), when $x>1$, the gradient is 0. If the contribution is defined as the product of the gradient and the input, the contribution of $x$ in this case is also 0, which clearly violates Sensitivity\cite{61}.
  
  \subsubsection{LayerCAM}

  Here, we take an image classification model as an example to introduce LayerCAM. Formally, let $f$ denote the image classifier, $\theta$ represents its parameters, and $y$ represents the predicted score. For a given image $I$ and target class \(c\),
  
  \begin{equation}\label{eq:eq2}
    y^c = f^c(I,\theta) 
  \end{equation}
  
  Let $A$ denote intermediate feature maps of a certain layer, $A^k$ denote the $k$-th feature map (the $k$-th feature dimension), \(g_{ij}^{kc} = \frac{\partial y^c}{\partial A_{ij}^k}\) represent the gradient of \(y^c\) with respect to the spatial location \((i,j)\) in \(A^k\), $w$ represent the weight of the feature map.
  
  \begin{equation}
    \label{eq:LayerCAM1}
    w_{ij}^{kc} = \text{relu}(g_{ij}^{kc})
    \end{equation}
  
  $\hat{A}$ represents the class activation map.
  \begin{equation}
    \label{eq:LayerCAM2}
    \hat{A}_{ij}^{kc} = w_{ij}^{kc} \cdot A_{ij}^{k}
    \end{equation}
  
  $M$ represents the final results, which combinning $\hat{A}$ along the feature dimension.
  \begin{equation}
    \label{eq:LayerCAM3}
    M^{c} = \text{ReLU}\left(\sum_{k} \hat{A}^{kc}\right)
    \end{equation}
  
  \subsubsection{ODAM}\label{sec:ODAM}
  
  ODAM is specifically proposed for object detection models and is essentially an application form of LayerCAM in object detection models. Unlike image recognition models, object detection models can distinguish between different instances. For an instance $p$, the model outputs category information as well as spatial coordinate information of the instance.
  
    During the calculation process of ODAM, unless otherwise specified, the representation of variables remains the same as that in LayerCAM. For an instance $p$, a feature layer is selected, and its calculation process is identical to that in LayerCAM. In object detection models, such as DETR, each instance has five outputs: category information $y$, and bounding box coordinates $x_1$, $y_1$, $x_2$, $y_2$. 

  Driven by a misguided intuition, ODAM interprets the box coordinates as taking the values of $x_1$, $y_1$, $-x_2$, $-y_2$. This might be because, when the feature region is centered, the expansion or contraction of the feature region has opposite effects on $x_1$, $y_1$ and $x_2$, $y_2$. However, this line of thinking is incorrect, please refer to \Cref{sec:Debunking}.
  These outputs are uniformly represented as $Y$. Here, \(g_{ij}^{k} = \frac{\partial Y}{\partial A_{ij}^k}\), where $Y$ can be, depending on the context of interpretation, the score $y^c$ for a certain category $c$, or the coordinates of the bounding box $x_1$, $y_1$, $-x_2$, $-y_2$. ODAM's computation process for interpreting a single output is almost identical to that of LayerCAM, with the only difference being the removal of the $relu$ in \Cref{eq:LayerCAM1}.

  ODAM provides a scheme that integrates the interpretation results from all outputs to produce the final interpretation (we do not endorse this scheme, refer to \Cref{sec:DETR}). In this context, $Y$ represents all outputs, and $\Phi$ denotes the operation of taking the maximum value of the interpretation results of all model outputs. (For specific details, please refer to the ODAM code: https://github.com/Cyang-Zhao/ODAM).
  
  \begin{equation}
    \label{eq:odam_core}
    \begin{split}
    M_{ij} &= \Phi \left( M_{ij}^{Y} \right)
    \end{split}
    \end{equation}
  
  However, we believe that merging the interpretation results of multiple different outputs can only bring about visual changes. The true interpretation still corresponds to a specific output.
  
  \subsubsection{DM}
  DM being an extension of \Cref{eq:odam_core} of ODAM. DM aims to integrate interpretation results from different outputs to obtain the final interpretation that better align with traditional evaluation metrics. We do not present the detailed computation process, as we oppose integrating different results as the final result and do not endorse traditional evaluation methods for model interpretation. We present DM primarily to illustrate that flawed evaluation metrics have misguided researchers' directions. LayerCAM was published in TIP 2021, ODAM was published in ICLR 2023 and TPAMI 2024, and DM was published in ICCVW 2025. These publications sufficiently demonstrate that gradient-based methods that disregard baselines have gained recognition from mainstream authoritative researchers in the field, while the flawed evaluation methods have not attracted any scrutiny. When something wrong is used long and widely enough, it becomes the standard. During some submission processes, we were required to adopt traditional evaluation methods for experimental comparisons. These issues constitute the significance of undertaking this work.

  \subsection{Integrated Gradients}
  
  The emergence of IG addresses the issue that gradients fail to satisfy Sensitivity\cite{61}, and IG satisfy Implementation Invariance\cite{61} at the same time.
  
  The naive principle of IG is to focus on every step of the input contribution's change from the input baseline to the current input, rather than merely concentrating on the final state. This approach helps avoid situations where gradients become ineffective in the case of $F(x)$ in \Cref{fig:relu}.
  
  IG is defined as the path intergral of the gradients along the straightline path from the baseline $x' \in R^{n}$ to the input $x \in R^{n}$. The integrated gradient along the $i_{th}$ dimension is defined as follows.
  \begin{equation}\label{eq:IG}
      \begin{aligned}
          &\text{IG}_i(x) \\
          & ::=F(x)-F(x')\\
          & ::=\int_{\alpha=0}^1\frac{\partial F(x^{\prime}+\alpha(x-x^{\prime}))}{\partial (x_i^{\prime}+\alpha(x_i-x_i^{\prime}))}  \frac {\partial (x_i^{\prime}+\alpha(x_i-x_i^{\prime}))}{\partial \alpha} d\alpha\\
          &=(x_i-x_i^{\prime})\int_{\alpha=0}^1\frac{\partial F(x^{\prime}+\alpha(x-x^{\prime}))}{\partial (x_i^{\prime}+\alpha(x_i-x_i^{\prime}))}d\alpha
      \end{aligned}
      \end{equation}
  
  The calculation process adopts a straightline path from the baseline input to the current input; for the advantages of this approach, please refer to the original paper.
  
  The integral of IG can be efficiently approximated via a summation (Riemann Sum). We simply sum the gradients at points occurring at sufficiently small intervals along the straightline path from the baseline $x'$ to the input $x$.

  \begin{equation} \label{eq:IG1}
    \begin{aligned}
      &\text{IG}_i^{approx}(x)\\
      &::=\frac{(x_i-x_i^{\prime})}{m}\times\Sigma_{k=1}^m\frac{\partial F(x^{\prime}+\frac{k}{m}(x-x^{\prime}))}{\partial (x_i^{\prime}+\frac{k}{m}(x_i-x_i^{\prime}))}
    \end{aligned}
  \end{equation}
  Here $m$ is the number of steps in the Riemman approximation of the integral. 

  \subsubsection{IG in Model Interpretation}
  TAM, BT, and DIX are model interpretation method based on IG, but all have been proven to contain significant errors\cite{cui2026IntegratedGradients}.
  
  \subsection{Unification of Gradient Methods and Integrated Gradient Methods}\label{sec:unificationgrad}
  
  In \Cref{eq:IG1}, we can find that the gradient method is actually a special case of IG when $m = 1$, and the baseline $x^{\prime}$ is all-zero.
  Formally, it simply performs a simple linear approximation of a complex nonlinear operation, as show in \Cref{eq:G}.
  \begin{equation} \label{eq:G}
    \begin{aligned}
      &\text{G}_i(x)\\
      &::=x_i\times\frac{\partial F(x)}{\partial (x_i)}
    \end{aligned}
  \end{equation}

  \subsection{Unification of Integrated Gradient Methods and Taylor Expansion}
  
  When approximating Integrated Gradients, if we start from the gradient at the baseline, that is, we also set the number of steps as $m$, but begin with $k = 0$,  
  \begin{equation} \label{eq:IG2}
    \begin{aligned}
      &\text{IG}_i^{approx}(x)\\
      &::=\frac{(x_i-x_i^{\prime})}{m}\times\Sigma_{k=0}^{m-1}\frac{\partial F(x^{\prime}+\frac{k}{m}(x-x^{\prime}))}{\partial (x_i^{\prime}+\frac{k}{m}(x_i-x_i^{\prime}))}
    \end{aligned}
  \end{equation}
  
  When $m = 1$, and the baseline $x^{\prime}$ is all-zero,
  \begin{equation} \label{eq:Taylor}
    \begin{aligned}
      &\text{IG}_i^{approx}(x)\\
      & ::=F(x)-F(x')\\
      &::=x_i\times\frac{\partial F(x^{\prime})}{\partial (x_i^{\prime})}
    \end{aligned}
  \end{equation}
  
  This is precisely the first-order term in the Taylor expansion. The Taylor expansion itself inherently features a well-defined baseline, representing a manifestation of baseline-oriented thinking. Taylor expansion also offers an approach for model interpretation. The issue, however, lies in the fact that for multivariate functions, Taylor expansions beyond the first order require the computation of a large number of mixed gradients, imposing excessively high computational demands and making implementation difficult. Therefore, we have abandoned this approach in our study. Our introduction of the relationship between Taylor expansion and Integrated Gradients here is primarily aimed at emphasizing that a baseline is essential in rigorous model interpretation work.

  \section{Our Method}

  Our method rigorously performs model interpretation with clear baseline based on IG. We correct the principles for baseline selection, and employ attribution error to rigorously evaluate the interpretation results.
  
  \subsection{Baseline Selection Principles}
  
  There are errors in the baseline selection of the original IG. The principles for selecting baselines in the original IG method are as follows:
  \begin{enumerate}
    \item IG recommend that developers check that the baseline has a near-zero score, this allows developers to interpret the output as a function of the input.
    \item IG would additionally like the baseline to convey a complete absence of signal, so that the features that are apparent from the attributions are properties only of the input, and not of the baseline.
  \end{enumerate}

  Regarding the first principle, it essentially involves selecting an input baseline based on the output to a certain extent. The second principle determines the baseline through the input. We have already elaborated on these in \Cref{sec:baseline}. The baselines for input and output are almost impossible to align. The reason why the IG paper proposed these two principles and managed to justify them is primarily because IG is mainly applied to object recognition models. In such models, the final output is the softmax output, which amplifies the relative differences among input values, allowing high-probability categories to dominate the probability output while compressing low-probability categories, making it relatively easy to achieve near-zero scores. However, when we apply this method to object detection models, where one of the outputs is the coordinates of the detection bounding box and thus lacks the softmax transformation, we find it challenging to implement the two baseline selection principles proposed in the IG paper. This can be seen in \Cref{fig:class,fig:box,fig:vgg}. When the input is a completely black image, the model's output bounding boxes for objects fail to meet the first principle of IG's baseline selection.The first principle for baseline selection in IG is actually incorrect, but this flaw may not be exposed in specific scenarios.

  Here, we correct the principles for baseline selection as follows:
  
  \begin{enumerate}
    \item The baseline for the entire model interpretation is solely determined by the input. 
    
    This differs from one of IG's baseline selection principles, as we completely disregard the output. For details, please refer to \Cref{sec:baseline}.
    \item The baseline can be chosen arbitrarily, but the final attribution is based on the difference between the input and the baseline.
    
    For instance, for an image input, the baseline can be any image, depending on the user's needs. However, selecting an all-zero image aligns most closely with how humans and models perceive images. Although this may seem similar to IG's second baseline selection principle, it actually offers more possibilities for baseline selection. Developers can compare the impact of differences between any two images on the output results according to their needs. Moreover, we do not approve of the situation where the IG paper appears to satisfy its first principle by choosing random noise images as baseline inputs. Such a scenario requires a rigorous interpretation approach and should not be directly treated as an interpretation for the outputs, refer to \Cref{sec:interpretation}.

  \end{enumerate}

  \subsection{Correct Interpretation of Model Interpretation Results}\label{sec:interpretation}
  
  We have introduced baselines into model interpretation through IG, and accordingly, the interpretation of model interpretation results requires adjustments. The correct approach to interpreting model interpretation results is as follows: our objective to interpret is the difference between the model's current output and the baseline output, the attribution results rely on the difference between the current input and the baseline input. As shown in \Cref{fig:class,fig:box,fig:vgg}, the object to interpret is the difference between the current output and the baseline output. The baseline output is determined by the baseline input. In our experiments, we selected a completely black image as the baseline input. Therefore, the attribution results rely on the difference between the current input and the completely black image, which essentially corresponds to the current input itself.

  \subsection{Accurate Evaluation of Model Interpretation Results}\label{sec:Evaluation}

  \subsubsection{Traditional Evaluation Methods}
  There are currently two main approaches to evaluate the performance of model interpretation methods: qualitative and quantitative. The qualitative approach involves comparing model interpretation results with human understanding of target features for consistency. There are two types of quantitative approaches. One is similar to qualitative evaluation, where the consistency between model interpretation results and feature regions is compared in a batch manner. The other involves perturbing the input according to model interpretation results—for example, removing corresponding elements in order of importance and observing the magnitude of change in the model output. The essence is to evaluate based on marginal effects.

  Consistency-based evaluation relies on a default assumption: that the way humans perceive things is the optimal solution, and that the model under test is already good enough to obtain this optimal solution. Under this assumption, comparing model interpretation results with human understanding of features—where higher consistency indicates a better interpretation method. First, we cannot be entirely certain that human cognition truly represents the optimal solution; the model may also learn other patterns on its own. Another issue is that we cannot guarantee the test model is sufficiently good. In that case, deviations in model interpretation results may stem either from the interpretation method itself or from the model's insufficient performance.
  
  The marginal-effect-based evaluation approach ignores data patterns. The information conveyed by data depends on the relationships among elements. The marginal-effect-based evaluation approach disrupts these data interdependencies and introduces uncertainty, making it difficult to ensure the accuracy and reliability of evaluation results. For instance, the artifact phenomenon \cite{DBLP:conf/ijcai/Xie0CZ23} confirms this problem. Readers are referred to the cited article for details on the artifact case. Regarding the limitations of marginal effects, we can illustrate this with a very simple everyday example. A man and a woman get married and have a child. When evaluated using marginal effects, if we assume there is no man in this relationship, then the child could not exist. Thus, the marginal-effect-based evaluation would attribute the entire contribution of having the child to the man. This is clearly incorrect. The deficiency of marginal effects was already identified as early as 1953\cite{1953A}, yet more than seventy years later, using marginal effects for result evaluation has become virtually an established gospel in the field of explainable AI.

  \subsubsection{Attribution Error}

  The introduction of IG has enabled more reasonable and accurate model interpretation. However, it is regrettable that the evaluation of model interpretation results in the original IG paper still relies on traditional methods. As shown in \Cref{fig:class,fig:box,fig:vgg}, we can precisely assess the quality of model interpretation results by the attribution error between the attribution results and the attribution targets, thereby eliminating the drawbacks of traditional evaluation methods. 

  Particularly, as shown in \Cref{fig:vgg}, when evaluated by traditional methods, the interpretability performance of our method at low stages appears to be very poor. However, from the perspective of attribution error, these interpretation results precisely reflect the contributions of features from that specific layer. This discrepancy arises from the varying degrees of feature extraction at different stages—the closer a layer is to the output, the more its features resemble semantic representations. This demonstrates that our approach of assessing interpretability through attribution error can circumvent the drawbacks inherent in traditional evaluation methods.
  
  \subsection{Model Interpretation through Features from different Layers}

  Feature-based model interpretation methods typically first select a feature layer for interpretation. For example, CAM and GradCAM rely on deep features—more precisely, the features from the last layer—while LayerCAM considers fusing interpretation results from different layers. The reason for this practice is that model interpretation results differ across feature layers. According to traditional evaluation methods, interpretation results derived from different layers vary in quality, so researchers usually select a specific feature layer for model interpretation.

Regarding the attribution error, we hold a completely opposite view. In the experiments shown in \Cref{fig:class,fig:box,fig:vgg}, we demonstrate that the attribution errors of interpretation results obtained from different feature layers are all very small. Under traditional evaluation methods, using different layers for interpretation yields results of varying quality, but under attribution error, there is no difference in using features from different layers for model interpretation. The hierarchical processing of data by the model is essentially a continuous process of feature extraction. Shallow feature integration is incomplete, so the resulting interpretation results intuitively appear poor. As the model goes deeper, feature extraction becomes increasingly refined and closer to semantics, which better aligns with human intuition—and according to traditional evaluation methods, this means the interpretation results are better. However, from a numerical perspective, we can treat the intermediate feature $A$ as the input and regard the subsequent network as a function $f'$, so the output is y = f'(A). When using IG for model interpretation, we can always accurately attribute the output to the input. IG brings an objective evaluation standard to model interpretation.

  LayerCAM attributes part of the reason for the differing interpretation results from features of different layers to the different spatial resolutions across layers. In \Cref{fig:class}, we present experimental results of our method based on ResNet features and Encoder features. Even when the two features have the same spatial resolution, the results still show that interpretation results based on deeper features better align with semantics. The essence of this phenomenon lies in the differing degrees of feature extraction across layers, rather than differences in spatial resolution.

  In short, features from any layer can sufficiently explain the model. The differences arise from the varying degrees of feature extraction at different depths, with no distinction in quality.

  In \Cref{sec:Evaluation}, we pointed out the limitations of traditional evaluation methods. The attribution error-based evaluation method we advocate is more objective. When the attribution error is sufficiently small, we have sufficient reason to believe that the model interpretation method is effective. If the interpretation results do not align with the feature regions recognized by humans, the reason may lie in the model's own performance being insufficient, or the model having learned a cognitive pattern different from that of humans, or the feature layer relied upon representing a feature extraction stage that still has limitations.

However, there is a hidden assumption here: we assume that model interpretation should reflect the model's complete extraction of data features, such as the understanding of image semantic content. We can attribute the model output to intermediate features at any layer, which is reasonable, but not every layer's features represent the model's complete extraction of semantic features. The widely adopted qualitative and quantitative experiments are actually built on the basis of human understanding of image semantics. Our human semantic features are the final features we have fully extracted. However, some methods perform model interpretation based on shallow features of the model. For instance, ODAM's interpretation of the DETR model is based on features extracted by ResNet rather than the features extracted by the final Transformer, which is unreasonable. The model's complete extraction of image semantic features lies in the features fed into the discriminator, not the intermediate features. Therefore, to fully reflect the features upon which the entire model makes inferences, selecting deep features is necessary. Consequently, ODAM's choice of using ResNet-output features when explaining DETR is inappropriate.

Additionally, the experimental results verify another phenomenon: the closer the features are to the output, the lower the degree of nonlinearity between the features and the output, and thus the attribution error is more likely to be smaller under the same conditions. We will provide a detailed analysis in the subsequent experiments.

\subsection{Method}\label{sec:method}
  We follow the parameter definitions used in LayerCAM and ODAM, and the computational process of our final method is as follows:
  
  \begin{equation} \label{eq:myIG}
    \begin{aligned}
      &\text{M}_{ij}^{k}(A,{A}')\\
      &::=\frac{(A_{ij}^k-{A_{ij}^k}')}{m}\times\Sigma_{k=1}^m\frac{\partial F({A^k}'+\frac{k}{m}(A^k-{A^k}'))}{\partial ({A_{ij}^k}'+\frac{k}{m}(A_{ij}^k-{A_{ij}^k}'))}
    \end{aligned}
  \end{equation}

    \begin{equation}
      \label{eq:LayerCAM3_2}
      {M}_{ij} = \sum_{k} \text{M}_{ij}^{k}
      \end{equation}
  
  Here $m$ is the number of steps in the Riemman approximation of the integral.

  To emphasize the flexibility in baseline selection, we express our process as $\text{M}_{ij}^{k}(A,{A}')$. Here, $A$ represents an intermediate feature from a certain layer, and ${A}'$ denotes the baseline for that intermediate feature. Using the intermediate feature $A$ as a demarcation, we divide the entire network into two parts: $f_1$ and $f_2$, where $A = f_1(x)$ and $y = f_2(A)$. The baseline is determined as ${A}' = f_1(x')$, and the output baseline is $y' = f_2({A}')$. The input baseline $x'$ can be arbitrarily defined within a reasonable range of the input. The final result is an attribution interpretation for $y - y'$ with respect to $x - x'$. We allow $x'$ to be arbitrarily defined within the reasonable input range, but when we only explain $x$, we still define $x'$ as an all-zero vector. The all-zero baseline is also adopted in subsequent experiments.

  It is particularly worth emphasizing here that, although ODAM can be regarded as a single-step IG, ODAM does not have an explicit input baseline. For example, when interpreting DETR, ODAM is implemented based on features output by ResNet, meaning that ODAM's baseline is an all-zero feature map with exactly the same shape as ResNet's output features, i.e., the baseline is $A'=0$. However, in this case, we cannot confirm a clear and unique input baseline $x'$ via ${A}' = f_1(x')$. This is a hidden drawback of ODAM: ODAM can never perform precise attribution explanation with baseline interpretation method (\Cref{sec:interpretation}).

  \section{Experiments}
  
  After unifying the gradient method and the IG method within the same framework, we can find that the gradient method is actually a single-step IG. In fact, there is no longer a need to compare these two methods in terms of interpretation accuracy, as IG method is bound to be more precise than the gradient method. Consequently, there is little need to conduct extensive experiments to validate the performance of these methods, as this would only waste computing power and energy.
  
  The main purpose of our subsequent experiments is not to compare the performance of ODAM and our method, but to demonstrate how to implement our method and how to correctly interpret model interpretation results using a baseline-oriented mindset. 
  
  We select the DETR model and the VGG model from the ODAM and LayerCAM papers for experimental demonstration.

  The environmental configurations and code for our experiments will be made publicly available. To replicate these experiments, it is preferable to have a GPU, as our experiments were conducted on an NVIDIA A100 GPU.

  The version of ODAM published in TPAMI has added some application scenarios for the model's interpretability results compared to its ICLR version, but this is not our focus. Here, we only discuss and compare the methods themselves, while the applications of the model's interpretability results are beyond the scope of our discussion.

  Additionally, it should be noted that ODAM has certain defects and even errors (\Cref{sec:DETR}), and the application of ODAM's results in the original paper also needs to be re-examined.
  
  To better present the results of model interpretation, we choose a more reasonable approach when creating visualizations. We denote the interpretation reslut as $S$, and then normalize $S$ using $\frac{S}{max(abs(S))}$, which can maintain a linear relationship between scores and preserve their positive and negative attributes. As shown in \Cref{fig:class},  red represents positive contribution, blue represents negative contribution, and white represents no contribution. A color bar is used for matching color and value. Colors of an image represent relative value within a single image, and cannot be used for comparing numerical values between images. Other images in this article follow these rules.
  In some figures, like \Cref{fig:duibi1}, in order to better demonstrate the allocation of overall contribution, we use the heatmap as the bottom layer, and then overlay the original image with a certain transparency on the heatmap layer for indication. 
  
  In all experiments based on the baseline analysis method advocated in this paper, we annotate the baseline detection box (black) and the model's output detection box under the current input (green), along with the corresponding baseline analysis data beneath each case image. When interpreting a target box, we use cyan to highlight the specific edge being explained, as shown in \Cref{fig:box}.

  \subsection{Interpretation of DETR}\label{sec:DETR}

  \subsubsection{Debunking ODAM}\label{sec:Debunking}
  
  Among the methods capable of providing model interpretation for DETR, the process-based method is GAE, while in feature-based methods, the ODAM paper demonstrates that ODAM outperforms other intermediate feature-based methods (such as GradCAM, GradCAM++\cite{DBLP:conf/wacv/ChattopadhyaySH18}, and D-RISE\cite{DBLP:conf/cvpr/PetsiukJMM0OS21}). GAE is also the primary method that ODAM paper compares with in the context of DETR interpretation, abbreviated as GAME in the ODAM paper. The ODAM authors evaluate the two as follows: ``The ODAM combo maps highlight information consistent with the Generic Attention-model Explainability (GAME)," and ``in contrast to GAME, ODAM is also able to highlight the important regions for predicting each coordinate of the bounding box."

  The following are our arguments refuting ODAM.
  
  \begin{itemize}
    \item GAE can similarly switch the target interpretation object from category output to coordinate output and is also capable of interpreting coordinate information, as shown in \Cref{{fig:gaezuobiao}}. 
    GAE is also a gradient method based on attention mapps from all layers, and the attention mapps is generated from features, meaning attention mapps and features are not independent , that is, the contribution of the input to the output is captured in both the attention maps and the intermediate features. All of these render baseline analysis methods inapplicable.   
  Therefore, Figure \Cref{fig:gaezuobiao} is solely used to refute the viewpoint of ODAM. GAE can switch its interpretation object to explain any output, and from this perspective, ODAM does not hold an advantage. Given that GAE cannot be analyzed using baseline methods, we refrain from evaluating its interpretation results. Moreover, since ODAM's interpretation of coordinates is incorrect (refer to next point), there is even no value in making a comparison.
  
\begin{figure}[!t]
    \centering
    \includegraphics[width=\columnwidth]{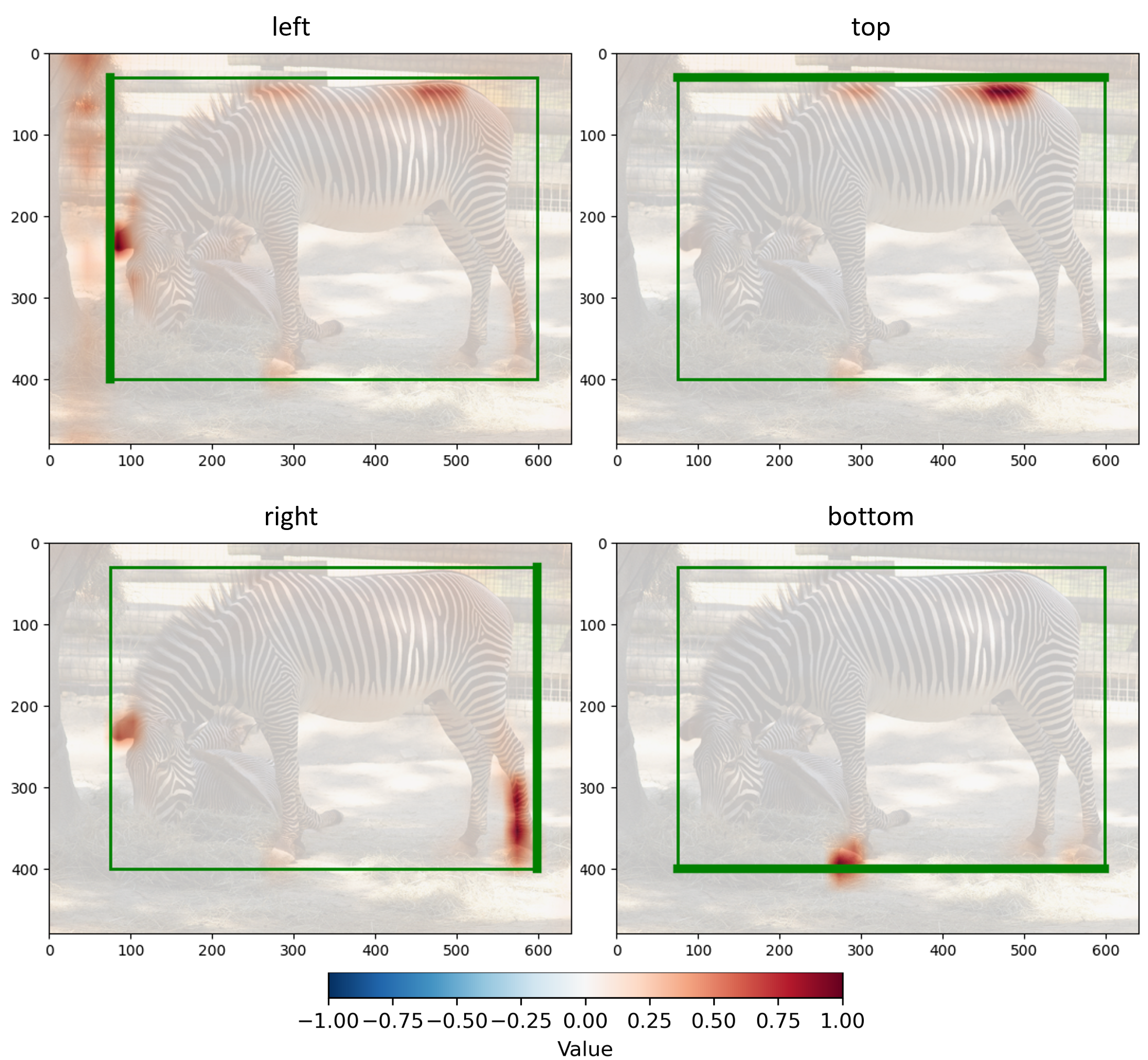} 
    \caption{GAE coordinate interpretation.}
    \label{fig:gaezuobiao}
  \end{figure}
  
  \item  ODAM's approach to interpreting coordinates is incorrect. Based on a flawed intuition, ODAM assigns the interpretation values for box coordinates as $x_1$, $y_1$, $-x_2$, $-y_2$. The reasoning behind this is that when considering the feature region as the center, the expansion or contraction of the feature region has opposite effects on $x_1$, $y_1$ and $x_2$, $y_2$. However, this line of thinking is erroneous. When viewed through the lens of baseline analysis methods, what we are actually interpreting is the difference between the current output and the baseline output. This seemingly reasonable idea, coupled with some misleading results, can easily lead to misunderstandings. As shown in \Cref{fig:box}, ODAM has a hidden baseline, and the interpretation results represent the difference between the current output and the baseline output.
  
  \item  It is clearly unreasonable to compare ODAM's interpretation results, which integrate both category and coordinate information, with GAE's pure category interpretation results. In \Cref{fig:duibi1}, we have re-presented the results focusing solely on category interpretation. Judging from the perspective of the qualitative evaluation criteria used in both the ODAM and GAE papers, ODAM's interpretation results fall significantly short of those of GAE. Please note that this evaluation is conducted using the qualitative evaluation criteria from the ODAM and GAE papers. Since baseline analysis methods cannot be applied to evaluate GAE's results, we refrain from making any assessments regarding the effectiveness of GAE's model interpretation.
  
\begin{figure}[!t]
    \centering
    \includegraphics[width=1\linewidth]{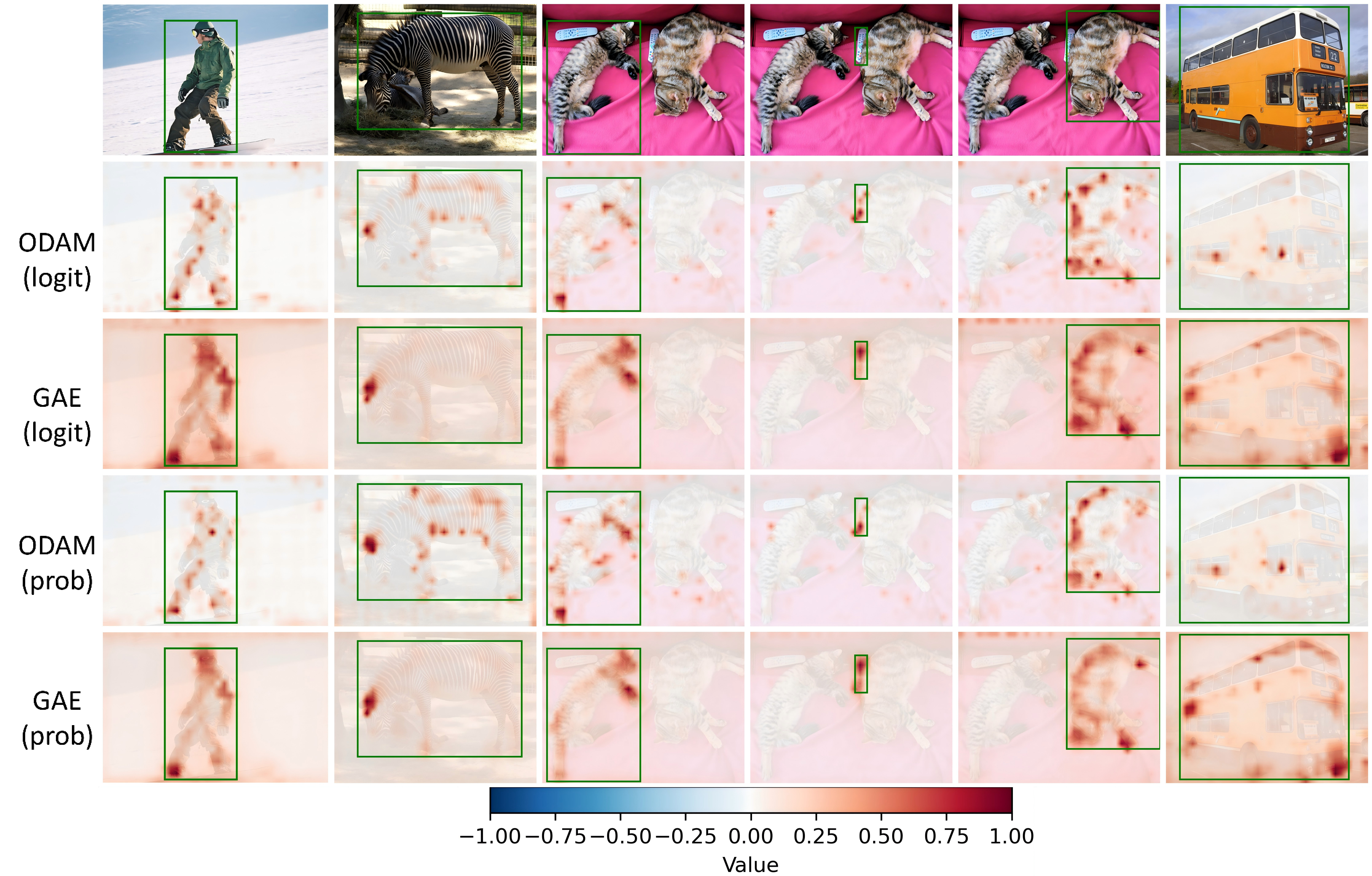} 
    \caption{Interpretation of category outputs (logits and probabilities) of the DETR\_demo model by ODAM and GAE. (ODAM strictly adheres to the original settings and provides interpretation based on the output features from ResNet.)}
    \label{fig:duibi1}
  \end{figure}
  
  \item  The experiment in ODAM\cite{ODAM2} employs DETR\_demo instead of the standard DETR model. The DETR\_demo was originally intended by the author's team as a simple demonstration of the model's underlying principles, and demo models are not used in formal research. For instance, the research in the original GAE paper utilizes the standard DETR model. Moreover, the team that proposed DETR has long since removed DETR\_demo-related content from their GitHub project. Judging from the evaluation criteria of ODAM and GAE, ODAM's results on DETR\_demo are inferior to those of GAE, and its results on the standard DETR model are even worse (\Cref{fig:duibi1,fig:duibi2}).
  
\begin{figure}[!t]
    \centering
    \includegraphics[width=1\linewidth]{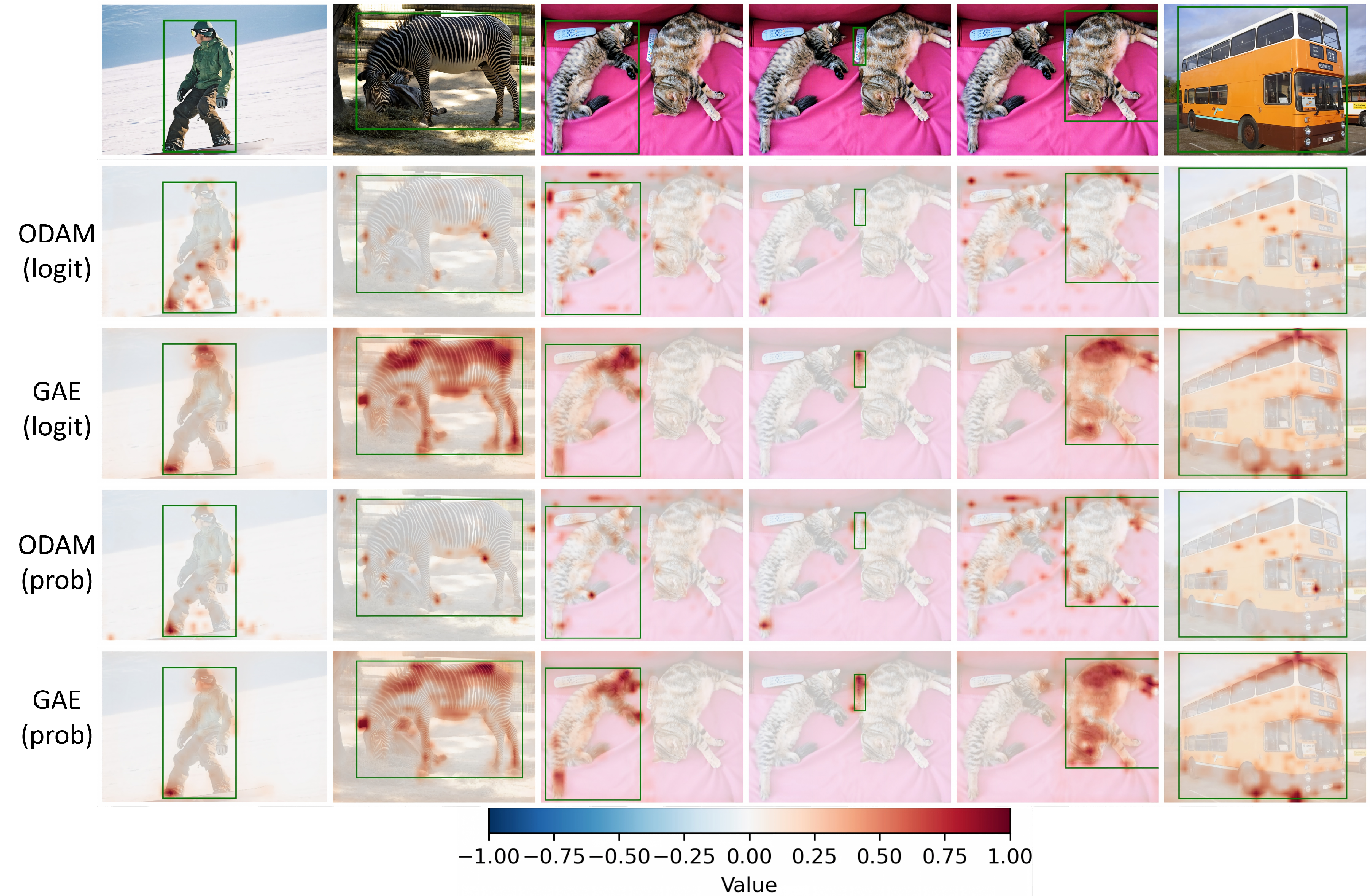} 
    \caption{Interpretation of category outputs (logits and probabilities) of the DETR model by ODAM and GAE. (ODAM strictly adheres to the original settings and provides interpretation based on the output features from ResNet.)}
    \label{fig:duibi2}
  \end{figure}

\end{itemize}

  \subsubsection{Comparative Experiments}\label{sec:comp}

  Since GAE is not applicable to the baseline analysis, or more specifically, the attribution error analysis, we only compare with ODAM in subsequent experiments.

  Based on the above refutations, we commence a comparison between ODAM and our method using the baseline analysis approach. It should be noted that we conduct experiments solely on the standard DETR model, as the DETR\_demo model has a simple structure and is merely for demonstration purposes; thus, we do not expend effort on the demo model. \textbf{We revise ODAM's method of interpreting coordinates, ensuring that the positive and negative attributes of coordinates are no longer altered.} In addition, we implement our method based on the output features of both ResNet and the encoder separately. Furthermore, combined with the case illustration of the VGG model in \Cref{sec:VGG}, we demonstrate that the closer the features are to the output, the more semantically relevant the feature extraction results are, and the closer the qualitative analysis results are to human cognition. However, it is important to note that features at any stage of the information processing can serve as a complete attribution for interpreting the output. Meanwhile, the closer the features are to the output, the lower the degree of nonlinearity and the higher the degree of linearity between the features and the output, resulting in fewer steps required when applying Integrated Gradients (IG) for model interpretation. This point becomes more evident in the VGG experiments in \Cref{sec:VGG}.

\begin{figure*}[!t]
    \centering
    \includegraphics[width=2\columnwidth]{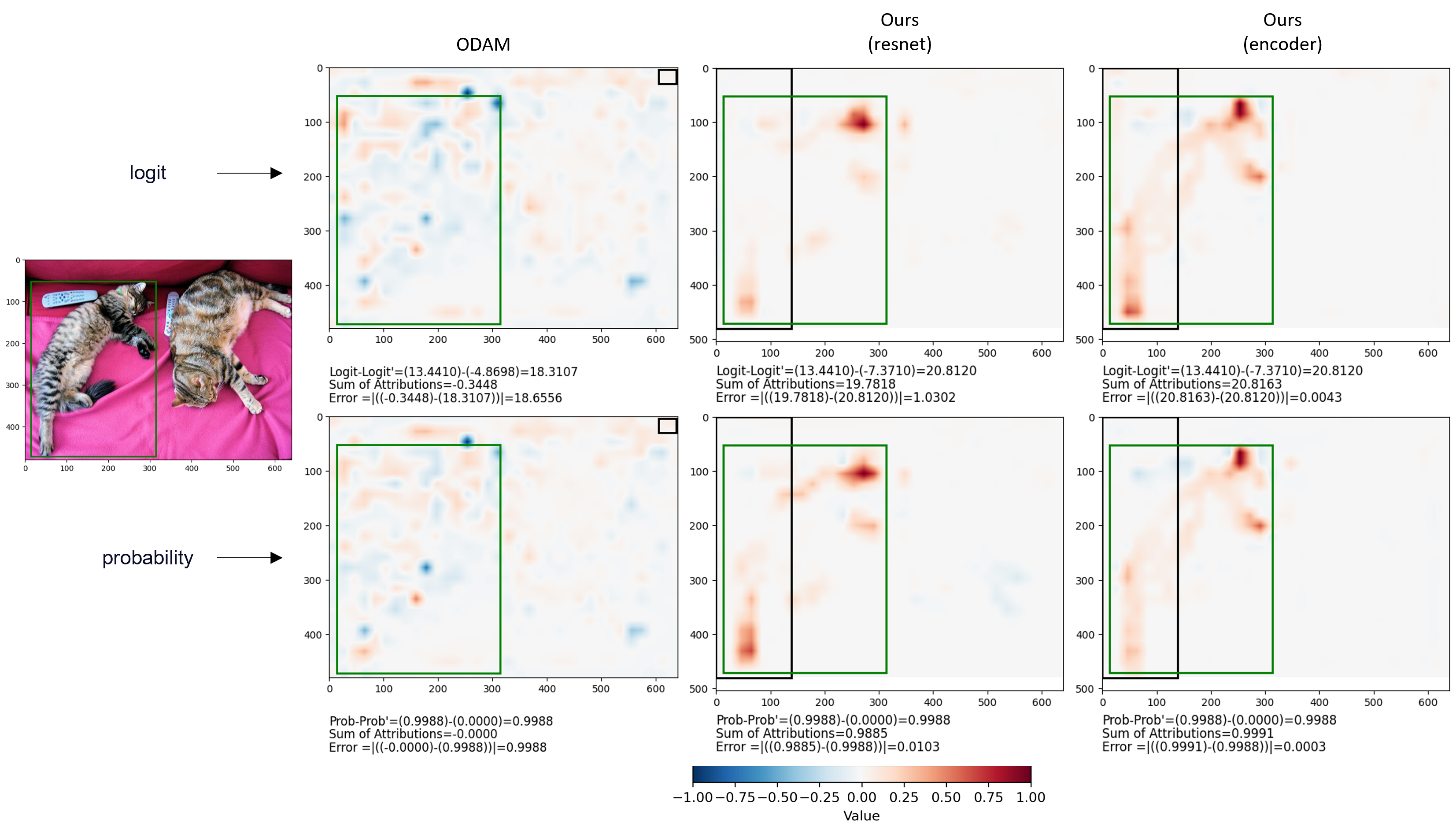} 
    \caption{Interpretation of category outputs (logits and probabilities) of the DETR modelby ODAM and our method.}
    \label{fig:class}
  \end{figure*}

\begin{figure*}[!t]
    \centering
    \includegraphics[width=2\columnwidth]{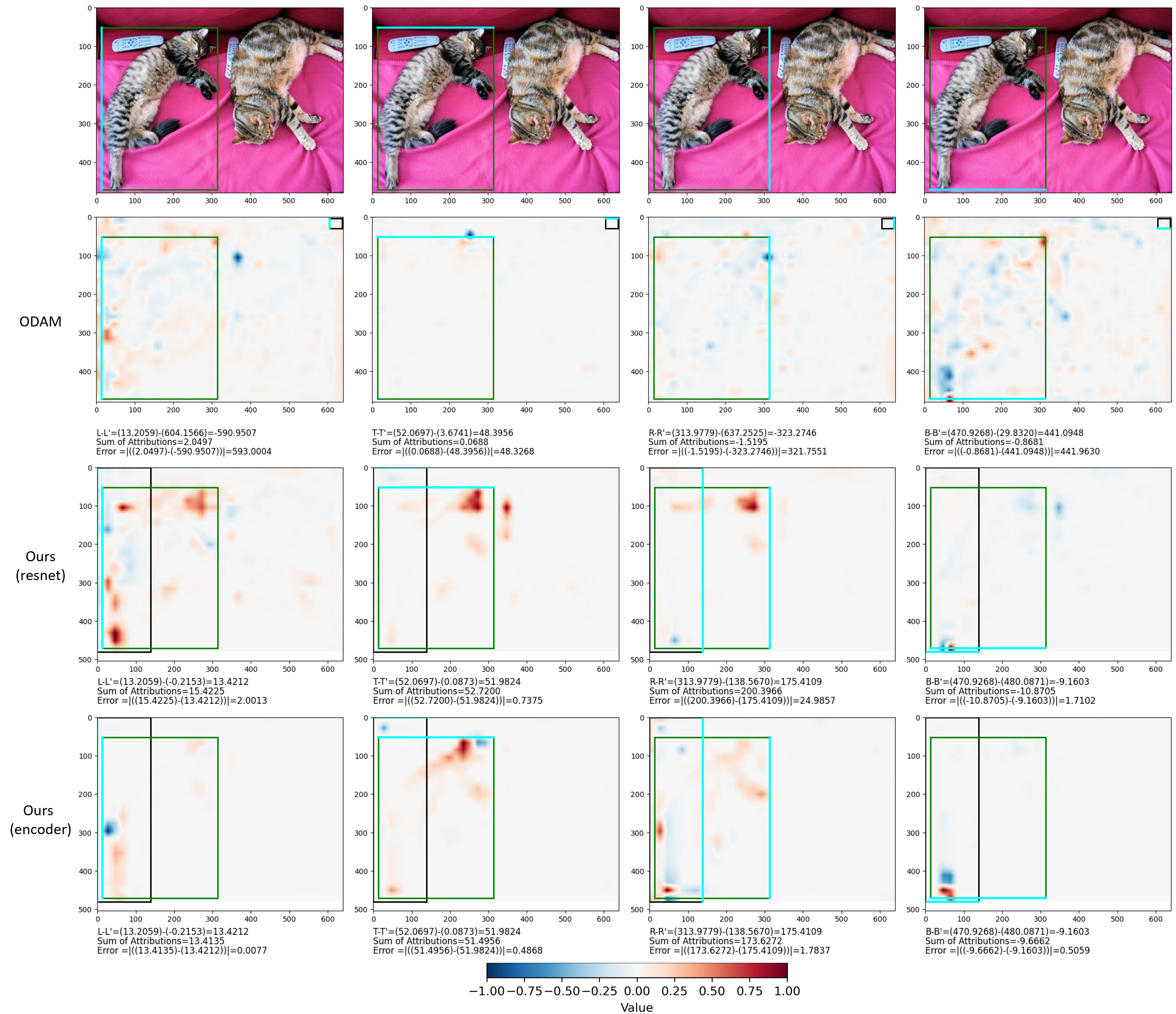} 
    \caption{Interpretation of coordinate outputs in the DETR model by ODAM and our method.}
    \label{fig:box}
  \end{figure*}

  Our interpretation results for categories are shown in \Cref{fig:class}, and those for bounding boxes are presented in \Cref{fig:box}. In all our experiments, the baselines are determined by all-zero inputs. Below each case illustration, we have indicated the difference between the output to be interpreted and the baseline output, the attribution results, the error. When interpreting categories, we display two sets of results: one using logit as the output for interpretation and the other using probability. In the figures, the green detection boxes represent the model's current output, while the black detection boxes denote the baseline output. We highlight the detection boxes being interpreted using cyan. In our experiments, we set the number of integration steps to 500. Additionally, we implemented our method based on features extracted from both the ResNet network output and the Encoder output.
  
  Upon a inspection of traditional evaluation methods, for category interpretation, the feature regions highlighted by our method align more closely with the correct feature regions. Upon a inspection of attribution error, our method is much better. Furthermore, it is noteworthy that, with the same number of integration steps, the interpretation accuracy based on Encoder features surpasses that based on ResNet features. This is because the closer the features are to the output being interpreted, the lower the degree of nonlinearity between the features and the output, and consequently, the fewer integration steps required. This point is more clearly demonstrated in \Cref{sec:VGG}.
  
  Additionally, it is particularly worth noting that, in the interpretation of detection boxes, our conclusion differs from that of ODAM. The highlighted conclusion in the original ODAM paper is that the model determines the corresponding coordinates of detection boxes based on partial regions or contours of the target's feature regions. In reality, the model's generation of detection box coordinates does not rely solely on the contours of feature regions, but rather on the combined effect of all elements. For example, in the interpretation results for the bottom border using our method implemented based on encoder features (as shown in the bottom-right case of \Cref{fig:box}), the model does not output coordinates based merely on the tips of the cat's paws; the upper parts of the cat's feet still play a very significant role. More precisely, every image element contributes to the final result, and the final output is the combined effect of all elements.
  
  We emphasize once again that there can be countless case demonstrations. The primary purpose of the case studies presented here is to illustrate the analytical process of our baseline-based method. Because ODAM is a single-step IG without a clear baseline input(please refer to \Cref{sec:method}).
  
  Although a performance comparison is not deemed necessary, we have designed batch experiments for further testing.
  
  We uniformly sampled 2,000 samples from 5,000 samples in the COCO dataset\cite{coco}. We set the batch size to 6, and if a batch contained images requiring resizing (stretching to alter their shapes), we discarded that batch. Ultimately, our batch experiments were conducted on a total of 1,194 samples. The reason for discarding images requiring resizing is that adapting the relevant code for such cases would be cumbersome and irrelevant to our experimental conclusions. We did not conduct the experiments on the entire COCO dataset due to the excessive computational time required. Instead, we employed uniform sampling, which does not affect the validity of our experimental conclusions. Our method consists of two stages: in the first stage, we set the number of integration steps to 256, and then selected samples with a relative error ($|\frac{error}{output-output_{baseline}}|$) greater than 0.5 for separate model interpretation with 2,000 integration steps. Finally, we compiled the experimental results and recorded the computational time. The results are presented in \Cref{tab:detrbatch}.

\begin{table}[!t]
  \caption{Batch experiments on DETR}
  \label{tab:detrbatch}
  \centering
  \resizebox{\linewidth}{!}{%
  \begin{tabular}{@{}lrrrrrr@{}}
  \toprule
  \multicolumn{1}{c}{} & \multicolumn{2}{c}{ODAM}                                                                                & \multicolumn{2}{c}{Ours (resnet)}                                                                       & \multicolumn{2}{c}{Ours (encoder)}                                                                      \\ \midrule
  \multicolumn{1}{c}{} & \multicolumn{1}{c}{Error} & \multicolumn{1}{c}{\begin{tabular}[c]{@{}c@{}}Runtime\\ (min)\end{tabular}} & \multicolumn{1}{c}{Error} & \multicolumn{1}{c}{\begin{tabular}[c]{@{}c@{}}Runtime\\ (min)\end{tabular}} & \multicolumn{1}{c}{Error} & \multicolumn{1}{c}{\begin{tabular}[c]{@{}c@{}}Runtime\\ (min)\end{tabular}} \\
  Class (Prob)         & 0.9834                    & 2.9643                                                                      & 0.2711                    & 168.5809                                                                    & 0.0286                    & 35.5282                                                                     \\
  Class (Logit)        & 15.8498                   & 2.8856                                                                      & 2.6590                    & 115.8652                                                                    & 0.5000                    & 29.9284                                                                     \\
  Box (Left)           & 344.9110                  & 2.8563                                                                      & 95.9336                   & 237.4421                                                                    & 29.2509                   & 78.2876                                                                     \\
  Box (Top)            & 160.3399                  & 3.0020                                                                      & 58.9888                   & 255.5348                                                                    & 19.0098                   & 68.0288                                                                     \\
  Box (Right)          & 176.7152                  & 2.8510                                                                      & 98.6844                   & 265.6704                                                                    & 30.0953                   & 77.6250                                                                     \\
  Box(Bottom)          & 341.4271                  & 2.9388                                                                      & 48.6525                   & 244.6687                                                                    & 14.9996                   & 73.7890                                                                     \\ \bottomrule
  \end{tabular}
  }
  \end{table}

  Please note that since ODAM always has an unknown input reference, it is inherently imprecise in principle. From this perspective, there is no need for us to analyze its model interpretation effectiveness. Purely from a mathematical standpoint, we can observe from \Cref{tab:detrbatch} that ODAM performs poorly. Given that ODAM and our method have different baselines, strictly speaking, it is not appropriate to compare the two methods. If we forcibly compare them based on the data attribution results, we find that in terms of error, our methods (ours (encoder) and ours (resnet)) perform the best, while ODAM performs the worst.

  \Cref{tab:detrbatch} shows that our method is much more accurate than ODAM, and it also verifies that the closer the features used for model interpretation are to the output being explained, the better the explanation accuracy under the same conditions. This is because the closer the features are to the output, the lower the degree of nonlinearity between the features and the output, and thus the higher the accuracy of IG.

Additionally, we emphasize again that the drawback of ODAM is not merely its poor performance shown in \Cref{tab:detrbatch}; more importantly, ODAM conceals a subtle yet fundamentally flawed weakness — ODAM lacks an explicit input baseline (refer to \Cref{sec:method}), which makes ODAM inherently imprecise from the perspective of baseline analysis.

  \subsection{Interpretation of VGG}\label{sec:VGG}
  
  We select VGG for experiments which is also used in LayerCAM paper. Since our interpretation within the DETR model already includes category interpretation, it is actually unnecessary to continue conducting experiments on the VGG model. However, the VGG network is relatively simpler and can more clearly reveal certain patterns.

  It is worth noting that, like ODAM, LayerCAM also lacks an explicit input baseline, which is an easily overlooked drawback.
  
\begin{figure*}[!t]
    \centering
    \includegraphics[width=0.9\linewidth]{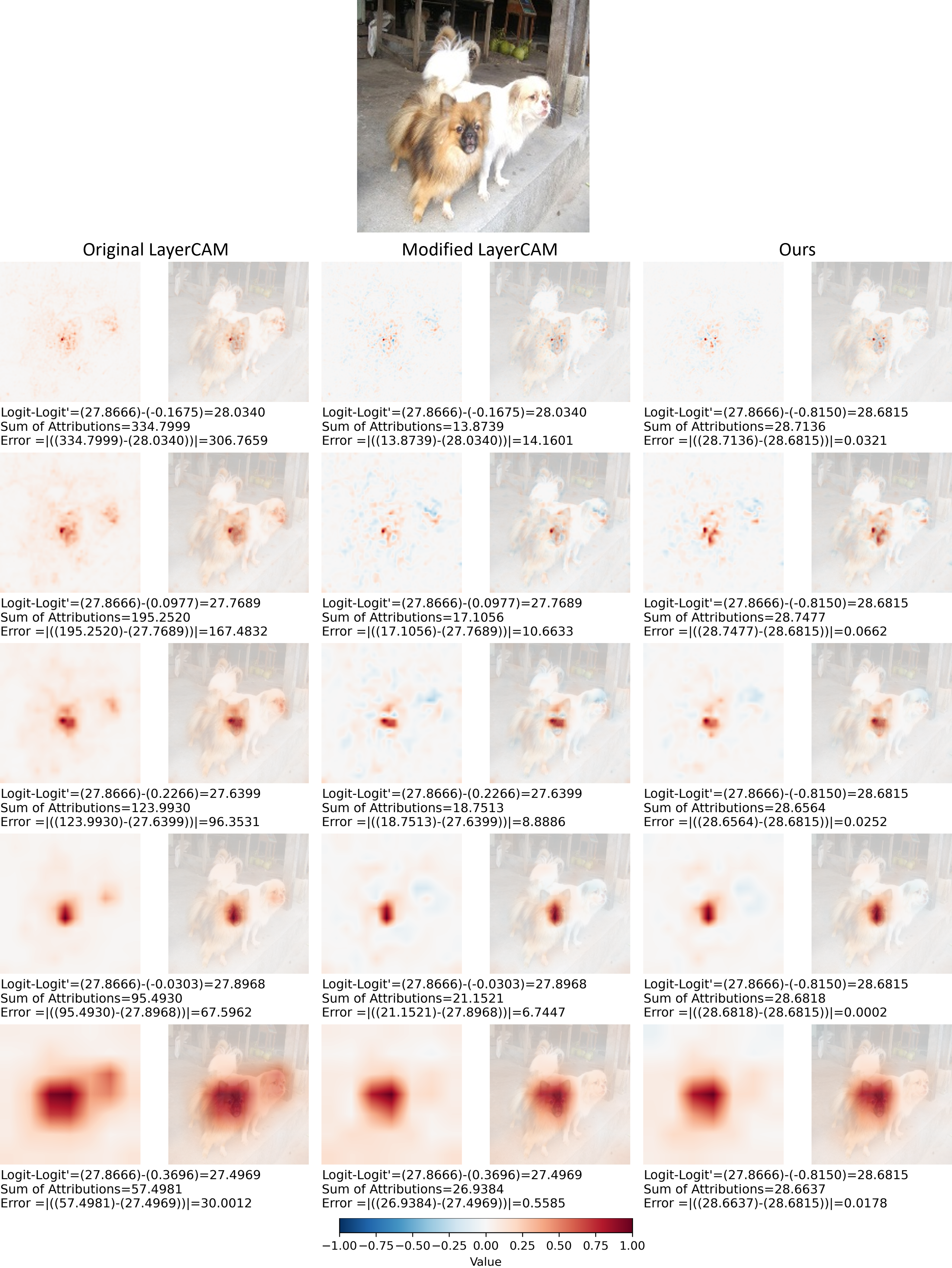} 
    \caption{VGG category logits interpretation.}
    \label{fig:vgg}
  \end{figure*}

  In the calculation process of LayerCAM, only the positive values of gradients are taken, which further increases the error in the interpretation results. Therefore, we present the original LayerCAM results, along with the interpretation results of a modified LayerCAM model that retains negative gradient values, as well as the results of our method. The experiments were designed in accordance with the original LayerCAM paper and were conducted in five stages. The experimental case demonstrations are shown in \Cref{fig:vgg}. Similarly, we uniformly sampled 5,000 ImageNet images for batch experiments, and the experimental results are presented in \Cref{tab:vgg}.
  
  In the experiments depicted in \Cref{fig:vgg}, the number of integration steps for our method was set to 500. In the batch experiments presented in \Cref{tab:vgg}, the number of integration steps was set to 256. Unlike the experiments in DETR, we no longer perform integration with more steps for samples with larger errors in this context.
  
  Upon observing \Cref{fig:vgg}, it can be seen that the modified LayerCAM exhibits smaller errors, with our method achieving the smallest errors. Although the interpretation results of the modified version appear visually similar to our method, our approach outperforms it in terms of error reduction. Additionally, it is crucial to note that neither the original LayerCAM nor the modified LayerCAM has a definitive baseline image, rendering the interpretation results of both methods less rigorous. 
  Furthermore, it can be observed that the model interpretation results based on deep-level features yield smaller errors. We believe this is because features closer to the output exhibit lower degrees of nonlinearity between the features and the outputs, thereby requiring fewer integration steps for IG and reducing the error of single-step IG. 
  This manifests as a gradual reduction in errors in the two special cases of single-step integration (original LayerCAM and modified LayerCAM).
  This observation is largely validated in batch experiments, with the only imperfection being that our method does not support this conclusion based on the results from the fourth and fifth stages. This may require further exploration, and what we present here is merely a possible conjecture.
  
    \begin{table}[!t]
      \centering
  \caption{Batch experiments on VGG}
  \label{tab:vgg}

  \resizebox{\linewidth}{!}{%

      \begin{tabular}{lrrrrrr}
      \hline
              & \multicolumn{2}{c}{Original LayerCAM}                                                                   & \multicolumn{2}{c}{Modified LayerCAM}                                                                   & \multicolumn{2}{c}{Ours}                                                                                \\ \hline
              & \multicolumn{1}{c}{Error} & \multicolumn{1}{c}{\begin{tabular}[c]{@{}c@{}}Runtime\\ (min)\end{tabular}} & \multicolumn{1}{c}{Error} & \multicolumn{1}{c}{\begin{tabular}[c]{@{}c@{}}Runtime\\ (min)\end{tabular}} & \multicolumn{1}{c}{Error} & \multicolumn{1}{c}{\begin{tabular}[c]{@{}c@{}}Runtime\\ (min)\end{tabular}} \\
      Stage 1 & 287.67038                 & 0.1870                                                                      & 9.277087                  & 0.3714                                                                      & 0.18437183                & 7.7305                                                                      \\
      Stage 2 & 164.09166                 & 0.1482                                                                      & 8.107285                  & 0.1535                                                                      & 0.10293762                & 5.5729                                                                      \\
      Stage 3 & 91.092766                 & 0.1491                                                                      & 7.3210387                 & 0.1501                                                                      & 0.08678595                & 3.3586                                                                      \\
      Stage 4 & 45.233196                 & 0.1545                                                                      & 5.9659696                 & 0.1507                                                                      & 0.07759022                & 1.5628                                                                      \\
      Stage 5 & 17.100424                 & 0.1529                                                                      & 0.8659283                 & 0.1548                                                                      & 0.13428052                & 0.8807                                                                      \\ \hline
      \end{tabular}
  }
      \end{table}

  Additionally, please note that while the interpretation results of the modified LayerCAM in the VGG model show comparable effectiveness to our method, its performance is significantly inferior when applied to the DETR model.

  \section{Limitations}
  
  Here we discuss a limitation of our baseline methods. We note that ODAM also conducted experiments on FCOS\cite{DBLP:conf/iccv/TianSCH19} and Faster R-CNN\cite{DBLP:conf/nips/RenHGS15}. Unlike DETR, these two models do not have stable outputs. The outputs of FCOS and Faster R-CNN are not fixed — the number of detected bounding boxes may vary across different images. However, Integrated Gradients requires a fixed output, as without stable outputs, there is no consistent integral. This is a limitation of applying the IG method. Nevertheless, this limitation is not entirely insurmountable. For instance, FCOS makes continuous predictions for each pixel, and Faster R-CNN produces fixed convolutional features. Interpreting these fixed features and, in turn, interpreting the model's inference may be a viable path forward.

  The focus of this study, as indicated by the title, is to emphasize the baseline that has been overlooked in the field of model interpretation, rather than to implement corresponding engineering applications on all models. Our future work will focus on applying our method to models without fixed outputs, such as FCOS and Faster R-CNN. It is important to note that ODAM, or LayerCAM, can be applied to such models, but the uncertain baseline renders them imprecise. From another perspective, LayerCAM and ODAM are essentially using instantaneous states to evaluate a process, or in other words, using single-step integration to approximate the complete integral. The stronger the nonlinearity of the network, the larger the error — these are unavoidable issues.

  \section{Conclusion}

  Our work emphasizes the long-overlooked baseline issue in the field of model interpretation. In this paper, we reformulate the task of model interpretation and the interpretation principles for model interpretation results to demonstrate the importance of the baseline.  We further unify gradient-based methods, Integrated Gradients (IG) methods, and Taylor expansion, clarifying the connections among them and explicitly identifying the baseline for each method. On this basis, we analyze the flaws and errors in related model interpretation methods (IG, LayerCAM, ODAM, Difference Map). We advocate evaluating the quality of model interpretation results precisely through the attribution error between the attribution result and the attribution target, rather than adopting flawed evaluation methods, such as those based on  marginal-effect or the assumption of perfect model performance. We revise IG and develope a model interpretation method with a clear and reasonable baseline, achieving better results. Our method supports model interpretation based on features from any layer. Interpretation based on features from different layers are all reasonable, and the differences among these results reflect varying degrees of feature extraction at different feature extraction stages.
  
  This work is not intended to target any researchers or their achievements; any reference to researchers or achievements is solely for the needs of academic research.

\bibliographystyle{IEEEtran}
\bibliography{danmotai}

\end{document}